\documentclass[letterpaper]{article} 
\usepackage{aaai25}  
\usepackage{times}  
\usepackage{helvet}  
\usepackage{courier}  
\usepackage[hyphens]{url}  
\usepackage{graphicx} 
\usepackage{natbib}  
\usepackage{caption} 
\usepackage{algorithm}
\usepackage{algorithmic}

\usepackage{newfloat}
\usepackage{listings}
\usepackage{bm}
\usepackage{amssymb}
\usepackage{multirow}
\usepackage{booktabs}
\usepackage{float}
\usepackage{makecell}
\usepackage{amsfonts}
\DeclareCaptionStyle{ruled}{labelfont=normalfont,labelsep=colon,strut=off} 
\floatstyle{ruled}
\newfloat{listing}{tb}{lst}{}
\floatname{listing}{Listing}
\title{Sign-IDD: Iconicity Disentangled Diffusion for Sign Language Production}
\author{
    Shengeng Tang, Jiayi He, Dan Guo, Yanyan Wei\thanks{Corresponding authors.}, Feng Li, Richang Hong*
}
\affiliations{
    School of Computer Science and Information Engineering, Hefei University of Technology\\

    \{tangsg, guodan, weiyy, fengli, hongrc\}@hfut.edu.cn, hejy4396@mail.hfut.edu.cn
}

\usepackage{bibentry}

\begin{document}

\maketitle

\begin{abstract}
Sign Language Production (SLP) aims to generate semantically consistent sign videos from textual statements, where the conversion from textual glosses to sign poses (G2P) is a crucial step. Existing G2P methods typically treat sign poses as discrete three-dimensional coordinates and directly fit them, which overlooks the relative positional relationships among joints. To this end, we provide a new perspective, constraining joint associations and gesture details by modeling the limb bones to improve the accuracy and naturalness of the generated poses. In this work, we propose a pioneering iconicity disentangled diffusion framework, termed Sign-IDD, specifically designed for SLP. Sign-IDD incorporates a novel Iconicity Disentanglement (ID) module to bridge the gap between relative positions among joints. The ID module disentangles the conventional 3D joint representation into a 4D bone representation, comprising the 3D spatial direction vector and 1D spatial distance vector between adjacent joints. Additionally, an Attribute Controllable Diffusion (ACD) module is introduced to further constrain joint associations, in which the attribute separation layer aims to separate the bone direction and length attributes, and the attribute control layer is designed to guide the pose generation by leveraging the above attributes. The ACD module utilizes the gloss embeddings as semantic conditions and finally generates sign poses from noise embeddings. Extensive experiments on PHOENIX14T and USTC-CSL datasets validate the effectiveness of our method. The code is available at: https://github.com/NaVi-start/Sign-IDD.

\end{abstract}

%

\begin{figure}[t]
  \centering
  \includegraphics[width=0.95\columnwidth]{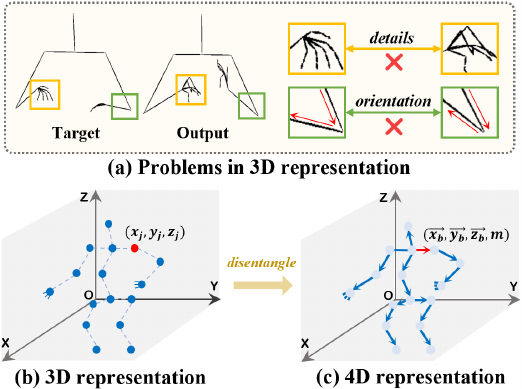}
  \caption{(a) Problems caused by using only traditional 3D representations for SLP. (b) Traditional 3D joint coordinate representation~\cite{saunders2020progressive,tang2022gloss}. (b) Proposed 4D bone representation. We take the neck joint as the root joint and define the parent-child joints for each bone along the skeletons. The Euclidean distance and direction vectors between parent-child joints are used to determine the length and orientation of the bones.
  }
  \label{fig:human_pose}
  \vspace{-5mm}
\end{figure}

\section{Introduction}
Sign Language Production (SLP) plays a crucial role in bridging the communication gap between the deaf and the general population, promoting inclusion and accessibility. This task is technically closely related to areas such as visual understanding~\cite{wei2024leveraging,li2024enhanced,guo2024shaping,li2024srconvnet} and cross-media reasoning~\cite{song2024emotional,wu2024intermediary,song2023emotion}. Given a textual description, SLP aims to transform it into the corresponding sequence of continuous signs automatically. These sequences can manifest as sign language poses~\cite{Saunders_Camgoz_Bowden_2020_A,saunders2020progressive} or sign language videos~\cite{saunders2022signing}. Currently, direct sign language video generation from spoken sentences remains a challenge due to the huge gap between sign vision and linguistics. Previous works usually translate spoken language into gloss\footnote{Glosses refer to minimal lexical items that match the meaning of signs in linguistics.} sequence (T2G) first and then generate sign pose video (G2P) based on gloss sequence~\cite{Saunders_Camgoz_Bowden_2020_A,saunders2020progressive}. Finally, the produced gesture poses are selectively used to generate realistic gesture videos~\cite{saunders2022signing}. Since T2G can be well addressed by Neural Machine Translation (NMT, language-to-language) based~\cite{othman2011statistical} and rule-based approaches~\cite{moryossef2021data}, G2P remains the key procedure for this task at this stage and is the focus of this work. 

Depending on the decoding strategy, current G2P methods are typically classified as either autoregressive~\cite{Saunders_Camgoz_Bowden_2020_A,saunders2020progressive,tang2022gloss} or non-autoregressive~\cite{huang2021towards,xie2024g2p,tang2024GCDM}. These efforts have promoted the development of SLP tasks, especially G2P-DDM~\cite{xie2024g2p} and GCDM~\cite{tang2024GCDM} as representative diffusion-based solutions, which improve the accuracy of generating sign poses. However, existing methods typically treat sign poses as discrete three-dimensional coordinates and only focus on the regression prediction of joint coordinates. These solutions overlook the modeling of relative positional relationships among joints, which hinders the effective generation of sign poses. 
Our goal is to generate clear and accurate gestures, especially in terms of poses that affect the semantic expression of sign language. In the example of Figure~\ref{fig:human_pose}(a), most of the joints in the output are already close to the target distribution in spatial position. However, their relative positions are quite different from the facts, especially in the finger details that are susceptible to deviation (yellow box). In addition, the orientation of the generated limbs also shows unexpected deviations (green box). 

A better constraint of bone details and limb orientation should be established in the generated pose using relative positions among the joints. Although some methods can deal with skeletal constraints by employing graph models~\cite{saunders2022skeletal}, they do not consider the inherent skeletal supervision and lead to higher computational costs. 
Fortunately, in the field of pose estimation, some emerging works have considered and verified the positive impact of constraining human bone length and orientation directly on prediction performance~\cite{cai2024disentangled}. Inspired by these works, we introduce an iconicity disentanglement strategy to enhance the relative position association (including orientations and distances) among joints.

\begin{figure*}[tbh]
  \centering
  \includegraphics[width=0.99\textwidth]{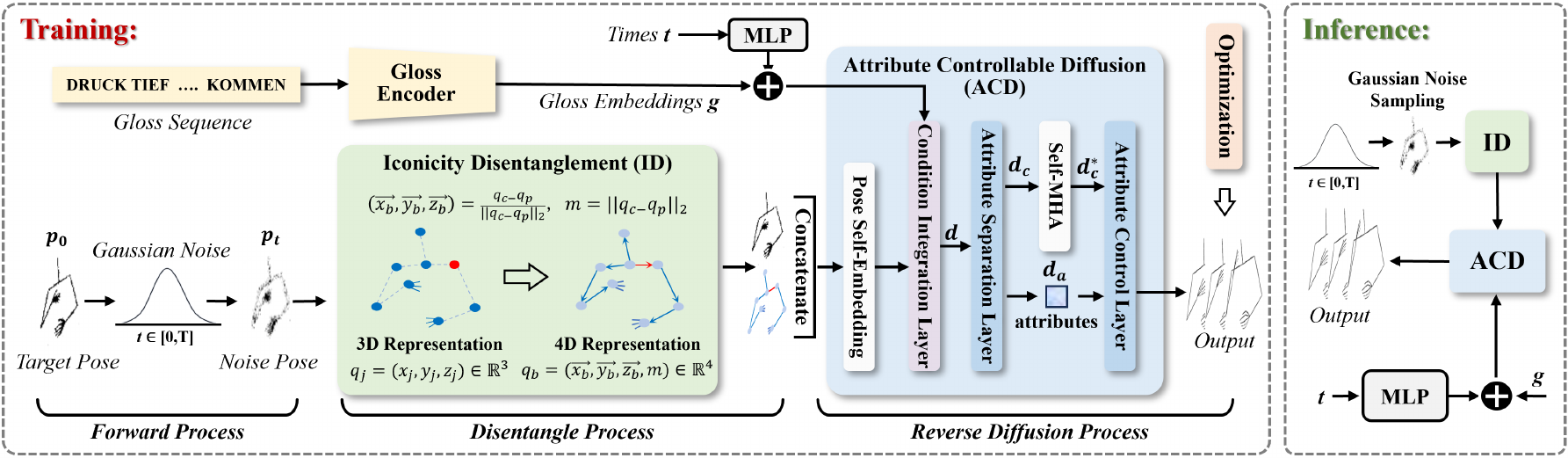}
  \caption{Overview of our framework. Given a gloss sequence, Sign-IDD generates a coherent sign pose video guided by gloss semantics. During training, we initially create Noise Pose $p_t$ by adding Gaussian noise to Target Pose $p_0$ for $t$ steps. Next, the 4D representation is derived from the 3D joint coordinates through Iconicity Disentanglement (ID). Then, the combination of the 3D and 4D representations is fed into the Attribute Controllable Diffusion (ACD) module, where gloss embeddings are integrated as a semantic condition. The attribute separation and control layers aim to separate skeletal attributes and guide pose generation. The final poses are optimized by applying joint and bone constraints. During inference, the initial $p_T$ is randomly sampled from Gaussian noise, with the disentanglement and reverse diffusion processes mirroring those used in training.  
  }
  \label{fig:Sign-IDD}
  \vspace{-5mm}
\end{figure*}

To this end, we propose a novel iconicity disentanglement diffusion framework, Sign-IDD, which aims to improve the expressiveness and accuracy of sign language gestures by enhancing the spatial association among joints. Sign-IDD incorporates a novel Iconicity Disentanglement (ID) module, to improve the perception of relative positions among joints. As shown in Figure~\ref{fig:human_pose}(b) and (c), two adjacent joints along each bone are regarded as parent-child joints. The ID module is designed to disentangle the conventional 3D joint representation into a 4D bone representation, comprising the 3D spatial direction vector and 1D spatial distance vector from parent to child joint. 
As shown in Figure~\ref{fig:Sign-IDD}, we further construct a diffusion-based SLP framework, where gloss embeddings are used as semantic conditions to guide sign language gesture generation. In this framework, the Attribute Controllable Diffusion (ACD) module is another core component, which further strengthens the association learning of joints in the generated poses. The ACD module first incorporates gloss conditions into the pose embeddings to achieve semantic guidance. The attribute separation layer aims to separate the bone direction and length attributes, and the attribute control layer is designed to optimize the sign generation under the guidance of the above attributes. Our main contributions are summarized as follows: 

\begin{itemize}
\item We innovatively introduce the concept of iconicity disentanglement, a novel strategy that transcends traditional joint coordinate regression fitting. Unlike most previous works that only adopt 3D joint coordinate representation, we utilize the disentangled 4D bone representation to further constrain the relative positions of joints, thereby ensuring the accuracy of sign pose details.

\item We further propose a novel diffusion-based gloss-to-pose SLP approach, containing an attribute controllable diffusion module controlling the orientation and length of bones, capable of generating accurate and robust 3D sign poses according to textual glosses. The introduced constraint of bones (\emph{i.e.}, $\mathcal L_{bone}$) further improves the quality of generated sign poses.

\item Exhaustive experiments on the PHOENIX14T and USTC-CSL datasets show that Sign-IDD significantly enhances pose accuracy, skeletal coherence, and linguistic fidelity, which outperforms SOTA methods.
\end{itemize}

\section{Related Work}
\subsection{Sign Language Production (SLP)}
Sign language research is a classic and hot topic in artificial intelligence. Early works focus on Sign Language Recognition (SLR) ~\cite{guo2021sign,Cui_Liu_Zhang_2017,Koller_2020,Guo_Zhou_Li_Wang_2018,Wang_Chai_Chen_2019} and Sign Language Translation (SLT)~\cite{tang2022graph,Camgoz_Hadfield_Koller_Ney_Bowden_2018,Cihan_Camgoz_Koller_Hadfield_Bowden_2020,Orbay_Akarun_2020,Guo_Tang_Wang}. 
Recently, increasing attention has been paid to Sign Language Production (SLP). 

Early SLP works translate sentences into sign representations using synthetic animation techniques~\cite{Mazumder_Mukhopadhyay_Namboodiri_Jawahar_2021,McDonald_Wolfe_Schnepp_Hochgesang_Jamrozik_Stumbo_Berke_Bialek_Thomas_2016,Segouat_2009}. These methods rely on rule-based lookups to pre-capture phrases, resulting in high collection costs and limited to predefined phrases. The development of deep models has sparked extensive research in SLP. 
Stoll \emph{et al.}~\cite{Stoll_Camgoz_Hadfield_Bowden_2020} adopt a multi-step process (i.e., text-to-gloss, gloss-to-pose, and pose-to-video) to generate sign video from the text. Saunders \emph{et al.}~\cite{saunders2020progressive} propose the first end-to-end SLP model to generate sign poses in an autoregressive manner. To improve the quality of generation, Saunders \emph{et al.}~\cite{Saunders_Camgoz_Bowden_2020_A} also introduces a multi-channel model with an adversarial discriminator. Mixture Density~\cite{saunders2021continuous} combines transformers and mixture density networks to model multi-modal continuous sequences. FS-NET~\cite{saunders2022signing} alleviates the error accumulation and the "mean sign pose" problem in the above autoregressive models. GEN-OBT~\cite{tang2022gloss} utilizes online reverse translation to enhance constraints of semantics. SignDiff~\cite{fang2023signdiff}, G2P-DDM~\cite{xie2024g2p}, and GCDM~\cite{tang2024GCDM} are both diffusion-based solutions that generate coordinate representations of sign poses from Gaussian noises. 
However, these methods treat sign poses as discrete 3D coordinates and overlook exploring relative positional associations among joints,  causing detail confusion and limb misorientation in generated poses. In contrast, our model employs an iconicity disentanglement strategy, separating joint coordinates into bone orientation and length, to better assist pose generation. 

\subsection{Diffusion Models}
Early generation works focus on Generative Adversarial Networks (GANs)~\cite{Goodfellow_Pouget-Abadie_Mirza_Xu_Warde-Farley_Ozair_Courville_Bengio_2017} and Variational AutoEncoders (VAEs)~\cite{kingma2013auto,makhzani2015adversarial}. 
Recently, diffusion models~\cite{Sohl-Dickstein_Weiss_Maheswaranathan_Ganguli_2015} have emerged as a novel approach garnering increasing attention, which have demonstrated remarkable achievements in various domains, including image generation~\cite{Ho_Jain_Abbeel_Berkeley,shang2024resdiff}, text generation~\cite{Li_Thickstun_Gulrajani_Liang_Hashimoto_2022}, speech synthesis~\cite{huang2022fastdiff}, and video generation~\cite{Ho_Salimans_Gritsenko_Chan_Norouzi_Fleet_2022}. 

The application of diffusion models in SLP is relatively rare and is still in its infancy. Baltatzis \emph{et al.} propose a diffusion-based model for generating motion sequences from textual transcriptions~\cite{Baltatzis_Potamias_Ververas_Sun_Deng_Zafeiriou_2023}. 
G2P-DDM~\cite{xie2024g2p} proposes the Pose-VQVAE framework, which combines VAEs and vector quantization to transform the continuous pose space generation into a discrete sequence generation problem. 
GCDM~\cite{tang2024GCDM} designs a gloss-driven conditional diffusion model and introduces a multi-hypothesis strategy to optimize sign pose generation.
Different from existing diffusion-based efforts~\cite{Baltatzis_Potamias_Ververas_Sun_Deng_Zafeiriou_2023,fang2023signdiff,xie2024g2p}, our Sign-IDD incorporates an attribute controllable diffusion module, to constrain joint associations by leveraging bone orientation and length attributes to guide the pose generation. This unique design enables the proposed solution to produce more accurate and controllable sign poses. 

\section{Preliminaries}
\subsection{Gloss to Pose Production}
Gloss to Pose production (G2P) is a crucial step of the SLP task. Given a gloss sequence $\mathcal{G} = \left \{g_n \middle| n = 1, 2, \ldots, N \right\} $ with $N$ glosses, G2P aims to transform it into a semantically consistent sign pose video $\mathcal{P} = \left\{p_s \middle| s = 1, 2, \ldots, S \right\}$ with $S$ frames. 
The goal of G2P is to learn a mapping function $\bm{F}(\mathcal{P} | \mathcal{G})$
that represents the probability distribution of generating a pose video $\mathcal{P}$ based on the given gloss sequence $\mathcal{G}$. The progressive generation process can be formalized as:
\begin{eqnarray}
    \bm{F}(\mathcal{P} | \mathcal{G}) = \prod_{s=1}^{S} \bm{F}(p_s | p_{<s}, \mathcal{G}).
\end{eqnarray}

\subsection{3D to 4D Representation} 
In existing works~\cite{saunders2020progressive,tang2022gloss,xie2024g2p}, sign poses are typically represented using a set of discrete 3D joint coordinates, which is widely adopted due to simplicity and computational efficiency. Each pose ${p}$ in pose sequence ${\mathcal{P}}$  is defined as a collection of joint points:
\begin{eqnarray}
    {p} = \left\{ q^{joint}_j = \left( {{x}_j}, {{y}_j}, {{z}_j} \right) \in \mathbb{R}^3 \middle| {j} = 1, 2, \cdots, J \right\},
\end{eqnarray}
where ${x_j}$, ${y_j}$ and ${z_j}$ represent the Cartesian coordinates of the $j$-th joint in 3D space, and $J$ is the total number of joints. 

However, these methods only consider the absolute positions of joints in 3D space and overlook the inherent constraints of human skeletal structure. This results in inaccuracies in pose generation, particularly with complex motions or subtle interactions between different body parts. To address these limitations, we introduce 4D bone representation, where each pose ${p}$ is expressed as a collection of bones in 4D form: 
\begin{eqnarray}
    {p} = \left\{ q^{bone}_b=\left( {\mathop{{x}_b}\limits ^{\rightarrow}}, {\mathop{{y}_b}\limits ^{\rightarrow}}, {\mathop{{z}_b}\limits ^{\rightarrow}}, {m} \right) \in \mathbb{R}^4 \middle| {b} = 1, 2, \cdots, B \right\},
\end{eqnarray}
where $B$ denotes the number of bones. Compared to the 3D joint representation, the 4D bone representation captures the correlations among joints and transforms absolute positions into relative ones, offering a more comprehensive and robust description of poses.

\section{Methodology}
Given a gloss sequence, our goal is to generate a sign pose video with consistent semantics, as illustrated in Figure~\ref{fig:Sign-IDD}. We first derive 4D bone representation from the original 3D coordinates through iconicity disentanglement (Sec. \textbf{Iconicity Disentanglement of Pose}). This 4D representation, combined with 3D coordinates, is then fed into the ACD module to enable skeletal attribute controllable pose generation guided by the gloss condition (Sec. \textbf{Attribute Controllable Diffusion}). Finally, the sign video is generated as a series of poses, optimized by applying joint and bone constraints (Sec. \textbf{Pose Generation and Optimization}). The following subsections provide detailed explanations.

\subsection{Iconicity Disentanglement of Pose}
\label{sec:ID}

In the previous section, we have explained the similarities and differences between 4D and 3D representations. Here, we will explain in detail how to obtain 4D bone representations from 3D joint coordinates. 
In the pose sequence ${\mathcal{P}}$, each pose ${p}$ corresponds to a series of 4D bone representations $\left( {\mathop{{x}_b}\limits ^{\rightarrow}}, {\mathop{{y}_b}\limits ^{\rightarrow}}, {\mathop{{z}_b}\limits ^{\rightarrow}}, {m} \right)$, which reflects the interconnections between adjacent joints. We regard the neck joint as the root node, whose 4D representation is especially noted as $({\mathop{0}\limits ^{\rightarrow}}, {\mathop{0}\limits ^{\rightarrow}}, {\mathop{0}\limits ^{\rightarrow}}, {0})$. We divide the parent ${q_{p}} \in \mathbb{R}^3$ and child ${q_{c}} \in \mathbb{R}^3$ joints based on the topology of the human body and using the root node as a reference, as shown in Figure~\ref{fig:human_pose}. Therefore, we can transform discrete 3D coordinates into 4D representations, converting absolute positions of joints into relative ones.   
Here, the direction and length of the bones can be obtained by calculating the directional vectors and Euclidean distances of adjacent joints in a three-dimensional space, which is called \textbf{Disentanglement}. The concept of \textbf{Iconicity} comes from linguistic semiotics~\cite{nielsen2021iconicity}, which means that although the obtained 4D representation is different from the original 3D representation in form, the pose semantics contained in them are essentially the same. 
The iconicity disentanglement from 3D to 4D is calculated as follows:
\begin{eqnarray}
    q^*=({\mathop{{x}_b}\limits ^{\rightarrow}}, {\mathop{{y}_b}\limits ^{\rightarrow}}, {\mathop{{z}_b}\limits ^{\rightarrow}}) = \frac{q_{c}-q_{p}}{{|| q_{c}-q_{p} ||}_2}, ~{m} = {|| q_{c}-q_{p} ||}_2, 
    \label{eq:AD}
\end{eqnarray}
where $b \in [1, B]$, $B=J-1$ denotes the number of bones. 

The advantage of 4D representation lies in its ability to model joint associations by converting absolute positions into relative ones. This captures intrinsic dynamic constraints of poses~\cite{cai2024disentangled}, enhancing robustness against anomalies in bone length and orientation, thereby reducing distortions and producing more accurate poses.

\begin{figure}[tbp]
\centering
\includegraphics[width=\columnwidth]{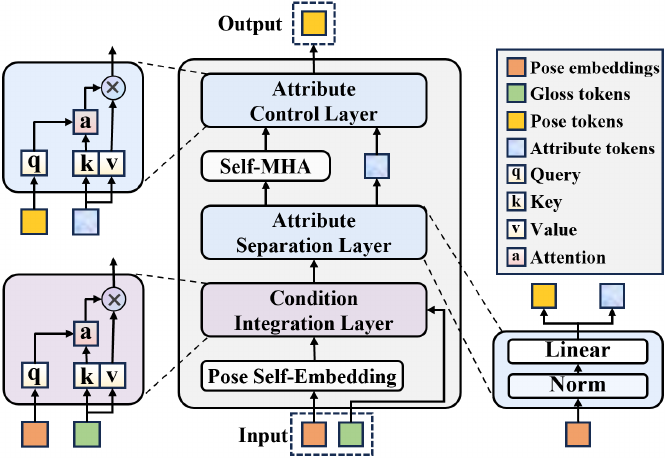}
\vspace{-3mm}
\caption{The main components of our ACD module.}
\vspace{-5mm}
\label{fig:ACD}
\end{figure}

\subsection{Attribute Controllable Diffusion}
Diffusion-based SLP involves two Markov chains: 1) a diffusion process that gradually introduces noise into the 3D poses, and 2) a reverse process that restores the original 3D poses from the 3D+4D noised poses through denoising. In the following sections, We will detail the forward and reverse processes of sign diffusion.

\subsubsection{Forward Process} As illustrated in Figure~\ref{fig:Sign-IDD}, the forward process in Sign-IDD begins by gradually infusing Gaussian noise $\epsilon \sim {\mathcal{N}} ({0}, {I})$ into the 3D pose ${p_0}$, increasing its intensity over time. This process is formulated as follows: 
\begin{eqnarray}
    \bm{Q}({p_t}|{p_0}):=\sqrt{\bar{a}_t} {p_0} + \epsilon\sqrt{1-\bar{a}_t},
    \label{eq:q}
\end{eqnarray}
where $\bar{a}_t := \prod_{t=0}^{T} a_t$ and $a_t := 1 - \beta_t$, and $\beta_t$ denotes the schedule for variance of cosine noise. When $T$ is sufficiently large, the distribution of $\bm{Q}({p_T})$ approaches an isotropic Gaussian distribution. Subsequently, ${p_t}$ undergoes the ID to transform into 4D representations of the form ${p_{t}'}$.

\subsubsection{Reverse Process}  In the training stage, ${p_t}$, ${{p_{t}'}}$ are concatenated into a fused representation $\tilde{p_{t}}$:
\begin{eqnarray}
    \tilde{p_{t}} = \{[{p_t},{p_{t}'}] \in \mathbb{R}^7\},
    \label{eq:merge}
\end{eqnarray}
and then, together with the textual semantics ${g}$ encoded by the Gloss Encoder~\cite{tang2024GCDM}, and the times of adding noise $t$, they are fed into the Attribute Controllable Diffusion (ACD) module $\mathcal{D}$ to restore the original unperturbed 3D poses:
\begin{eqnarray}
    {\tilde{p_0}} = \mathcal{D}(\tilde{p_{t}}, {g}, t).
    \label{eq: inference}
\end{eqnarray}

During inference, we initialize the 3D pose ${p_{T}}$ by sampling the noise from the unit Gaussian. As shown in Figure ~\ref{fig:Sign-IDD}, ${p_T}$ undergoes the disentanglement process to obtain its corresponding 4D representation ${p_T'}$. Subsequently, ${p_T}$ and ${p_T'}$ are merged into $\tilde{p_{T}}$, following Equation ~\ref{eq:merge}.

In the ACD module, we utilize the Multi-Head Attention (MHA) mechanism~\cite{vaswani2017attention} several times to achieve sequence self-embedding and multi-stream feature fusion. The MHA used in this work can be formulated as: 
\begin{eqnarray}
\left\{\begin{array}{l}
    {MHA}(Q, K, V) = [head_1, \cdots, head_h] \cdot W,\\
    head_i = Attention(Q W_i^Q, K W_i^K, V W_i^V), \\
    Attention(Q, K, V) = softmax(\frac{Q {K^T}}{\sqrt{d_k}})V,
\end{array}\right.
\end{eqnarray}
where $Q$, $K$, and $V$ represent the query, key, and value vectors respectively, $h$ denotes the number of heads, and $W_i^Q$, $W_i^K$, $W_i^V$, $W$ are learnable parameters. 

Figure~\ref{fig:ACD} illustrates the main components of the designed ACD module in our Sign-IDD. 
To refine the input pose sequence, we apply \textbf{\emph{pose self-embedding}} for self-encoding (SE) and positional encoding (PE), formulated as:
\begin{eqnarray}
    \hat{p_{t}} = SE(\tilde{p_{t}}) + PE(s),
\end{eqnarray}
where $SE$ is achieved through a linear embedding layer and $PE$ is implemented using a predefined sinusoidal function to encode the temporal information. 

The \textbf{\emph{condition integration layer}} aims to introduce gloss semantics to guide pose feature embedding, which is implemented based on MHA. The process of obtaining pose features $d$ with gloss semantics is expressed as: 
\begin{eqnarray}
    d=MHA(Q, K, V)|_{Q=\hat{p_t}, K=V=g}. 
\end{eqnarray}
For the fused pose features $d$ that already contain gloss semantics, we expect to separate the skeletal attributes (such as bone orientation and length) from them as supervisory cues to control the sign pose generation. Thus, we design an \textbf{\emph{attribute separation layer}} that reprojects the pose feature $d$ into the 7D space and separates the coordinate features $d_c \in \mathbb{R}^3$ and the attribute features $d_a \in \mathbb{R}^4$. This process is the inverse of Equation~\ref{eq:merge}. 
 
Next, the 3D coordinate features $d_c$ are passed through an independent MHA layer, yielding updated coordinate features $d^*_c$. Then, an MHA-based \textbf{\emph{attribute control layer}} integrates the attribute features $d_a$ into $d^*_c$, to refine and optimize skeletal details in sign poses. This process is formalized as: 
\begin{eqnarray}
    d_p=MHA(Q, K, V)|_{Q={d^*_c}, K=V={d_a}}. 
\end{eqnarray}

Finally, we obtain the 3D pose hypothesis $\tilde{p_0}$ from the noisy pose $\tilde{p_t}$ through ACD. This constitutes the input of $\mathcal{D}$ for the ensuing time step, expressed as
\begin{eqnarray}
    p_{t-1} = \sqrt{\bar{a}_{t-1}} \cdot \tilde{p_0} + \sqrt{1-\bar{a}_{t-1}-\sigma_t^2} \cdot \epsilon_t + \sigma_t\epsilon,
\end{eqnarray}
where $t$ and $t-1$ are the current time step and the next time step, respectively, and the initial $t = T$. $\epsilon \sim {\mathcal{N}} ({0}, {I})$ is a standard Gaussian noise independent of $p_0$ and 
\begin{eqnarray}
\left\{\begin{array}{l}
     \epsilon_t = ({{p_{t}}}-\sqrt{\bar{a}_{t}} \cdot {{\tilde{p_0}}}) / \sqrt{1-\bar{a}_{t}},\\
     \sigma_t = \sqrt{(1-\bar{a}_{t-1})/(1-\bar{a}_{t})} \cdot \sqrt{1-\bar{a}_{t}/\bar{a}_{t-1}},
\end{array}\right.
\end{eqnarray}
where $\epsilon_t$ is the noise at timestep t (derived from Equation ~\ref{eq:q}) and $\sigma_t$ controls the stochastic during the diffusion process.

In this stage, $p_{t-1}$ is utilized as input of $\mathcal{D}$ to regenerate and update ${p_0}$, which is repeated $I$ times. Initiated at $T$, the timestep for each iteration is computed as $T = T - (1 - i / I)$, where $i \in [0, I]$. The adjustable parameter $i$ controls the diversity and quality of the generated results.

\subsection{Pose Generation and Optimization}
\subsubsection{Pose Generation} 
In this part, we describe the process of generating the final poses $\hat{p_0}$ from the output feature $\tilde{p_0}$ of the ACD module. In practice, we employ a Multi-Layer Perception (MLP) to reproject $\tilde{p_0}$ to $\hat{p_0}$, represented as:
\begin{eqnarray}
    \hat{p_s} = MLP(LayerNorm(\tilde{p_0})). 
\end{eqnarray} 
Here, we optimize the sign pose generation process by constraining the joint coordinates (\emph{i.e.}, joint constraint) and the bone orientation (\emph{i.e.}, bone constraint).

\subsubsection{Joint Constraint} 
Following~\cite{huang2021towards,saunders2020progressive,saunders2021mixed,Viegas_Inan_Quandt_Alikhani}, we adopt a joint loss to constraint the accuracy of the joint positions in poses, ensuring precise matching with the ground truth. The joint constraint $\mathcal{L}_{joint}$ is defined as follows: 
\begin{eqnarray}
    \mathcal L_{joint} = \frac{1}{S}\sum_{s=1}^{S}|p_s-\hat{p_s}|,
\end{eqnarray}
where $p^s_0$ and $\hat{p^s_0}$ represent the ground-truth and generated 3D pose at the $s$-th frame, respectively.

\subsubsection{Bone Constraint}
To better depict complex motion details during training, we introduce $\mathcal L_{bone}$ to improve the accuracy of bone orientations in the generated poses: 
\begin{eqnarray}
    \mathcal L_{bone} = \frac{1}{S}\sum_{s=1}^{S}(q^*_s-\hat{q^*_s})^2,
\end{eqnarray}
where $q^*_s$ and $\hat{q^*_s}$ represent the bone orientations derived from $p^s_0$ and $\hat{p^s_0}$ according to Equation~\ref{eq:AD}. 
The overall objective is:
\begin{eqnarray}
    \mathcal{L} = \mathcal L_{joint} + \lambda \mathcal L_{bone}.
\end{eqnarray}

\section{Experiments}

\begin{table*}[!htbp]
\renewcommand\arraystretch{1.0}   
\caption{Performance comparison on PHOENIX14T. `\dag' indicates the model is tested by us under a fair setting. \textbf{NDBM}: \textbf{N}on-\textbf{D}iffusion \textbf{B}ased \textbf{M}ethods; \textbf{DBM}: \textbf{D}iffusion \textbf{B}ased \textbf{M}ethods.}
\vspace{-3mm}
\centering
\resizebox{\textwidth}{!}{
\begin{tabular}{lccccccccccccccc}
\toprule[1pt]
\multirow{2}{*}{Methods}&\multicolumn{7}{c}{DEV} & ~ &\multicolumn{7}{c}{TEST} \\
\cline{2-8}
\cline{10-16}
~ &B1$\uparrow$ &B4$\uparrow$ &ROUGE$\uparrow$ &WER$\downarrow$ &FID$\downarrow$ &MPJPE$\downarrow$ &MPJAE$\downarrow$& ~ &B1$\uparrow$ &B4$\uparrow$ &ROUGE$\uparrow$ &WER$\downarrow$ &FID$\downarrow$ &MPJPE$\downarrow$ &MPJAE$\downarrow$\\ 
\midrule[0.5pt]
\multicolumn{1}{l}{Ground Truth}  &29.77 &12.13 &29.60 &74.17&0.00&0.00&0.00 &~ &29.76 &11.93 &28.98 &71.94 &0.00&0.00&0.00\\ 
\midrule[0.5pt]
\multicolumn{16}{l}{\bf{\emph{NDBM}}}\\
\multicolumn{1}{l}{PT-base$^\dag$$_{ECCV 2020}$} &9.53 &0.72 &8.61 &98.53 &2.90&41.92&33.74& ~ &9.47 &0.59 &8.88 &98.36 &3.22&51.35&33.17\\
\multicolumn{1}{l}{PT-GN$^\dag$$_{ECCV 2020}$} &12.51 &3.88 &11.87 &96.85 &2.98&40.63&28.25& ~ &13.35&4.31&13.17&96.50&3.33&50.8&28.81 \\
\multicolumn{1}{l}{NAT-AT$_{MM 2021}$} &-- &-- &-- &-- &--&--&--& ~ &14.26 &5.53 &18.72 &88.15 &--&--&--\\
\multicolumn{1}{l}{NAT-EA$_{MM 2021}$} &-- &-- &-- &-- &--&--&--& ~ &15.12 &6.66 &19.43 &82.01&--&--&-- \\
\multicolumn{1}{l}{DET$_{*SEM 2023}$}  &17.25&5.32 &17.85&-- &--&--&--& ~ &17.18 &5.76 &17.64 &--&--&--&--\\
\multicolumn{1}{l}{GEN-OBT$_{MM 2022}$ } &24.92&8.68 &25.21&82.36&2.54&41.47&26.64 & ~ &23.08 &8.01 &23.49 &81.78&2.97&52.9&27.53\\ 
\midrule[0.5pt]
\multicolumn{16}{l}{\bf{\emph{DBM}}}\\
\multicolumn{1}{l}{D3DP-sign$^\dag$$_{ICCV 2023}$} &17.20 &5.01 &17.94 &91.51 &2.38 &39.42 &25.73& ~ &16.51&5.25 &17.55 &91.83 &2.63 &47.65 &25.92\\
\multicolumn{1}{l}{G2P-DDM$_{AAAI 2024}$} &-- &-- &-- &-- &-- &-- &--& ~ &16.11&7.50 &-- &77.26 &-- &-- &--\\
\multicolumn{1}{l}{GCDM$_{TOMM 2024}$} &22.88&7.64&23.35&82.81 &-- &-- &--& ~ &22.03&7.91&23.20&81.94 &-- &-- &--\\
\multicolumn{1}{l}{\textbf{Sign-IDD (Ours)}}
&\bf{25.40}&\bf{8.93}&\bf{27.60}&\bf{77.72} &\bf{2.22} &\bf{39.11} &\bf{25.34}& ~ &\bf{24.80}&\bf{9.08}&\bf{26.58}&\bf{76.66} &\bf{2.46} &\bf{47.19} &\bf{25.37}\\ 
\bottomrule[1pt]
\end{tabular}}
\label{tab:main1}
\vspace{-5mm}
\end{table*}

\begin{table}[tbp]
\renewcommand\arraystretch{1.1}   
\caption{Performance comparison on USTC-CSL.}
\vspace{-3mm}
\centering
\resizebox{\columnwidth}{!}{
\begin{tabular}{lcccccc}
\Xhline{1pt}
Methods&B1$\uparrow$ &WER$\downarrow$ &FID$\downarrow$ &MPJPE$\downarrow$ &MPJAE$\downarrow$\\ 
\Xhline{0.5pt}
Ground Truth                  &69.10 &47.38 &0.00&0.00&0.00\\ 
\Xhline{0.5pt}
PT-base$_{ECCV 2020}$    &22.32 &87.64 &0.49&175.14&21.93\\
PT-GN$_{ECCV 2020}$  &24.42&84.01&0.46 &103.44 &18.97 \\
GEN-OBT$_{MM 2022}$      &38.31 &70.50&0.41&78.98&12.86\\ 
D3DP-sign$_{ICCV 2023}$  &59.37&53.62&0.34&79.27&11.08\\
\Xhline{0.5pt}
\textbf{Sign-IDD (Ours)} &\bf{65.26}&\bf{50.15} &\bf{0.31} &\bf{72.20} &\bf{10.92}\\ 
\Xhline{1pt}
\end{tabular}}
\label{tab:main2}
\vspace{-5mm}
\end{table}

\subsection{Experimental Settings}
\subsubsection{Datasets} We evaluate the proposed method on two benchmarks: PHOENIX14T~\cite{Camgoz_Hadfield_Koller_Ney_Bowden_2018} and USTC-CSL~\cite{huang2018video}. PHOENIX14T consists of 8,257 instances featuring 2,887 unique German words and 1,066 signs, known for its complexity. USTC-CSL encompasses 100 Chinese sign language sentences performed by 50 signers and is divided into 4,000 training instances and 1,000 testing instances~\cite{guo2018hierarchical}.

\subsubsection{Evaluation Metrics} Following the existing works~\cite{huang2021towards,saunders2020progressive,saunders2021mixed,Viegas_Inan_Quandt_Alikhani}, a SLT model named NSLT~\cite{Camgoz_Hadfield_Koller_Ney_Bowden_2018} is employed to back-translate sign poses into textual glosses and compare them with references for calculating metrics such as BLEU, ROUGE, and WER. In addition, we also report the FID, Mean Per Joint Position Error (MPJPE), and Mean Per Joint Angle Error (MPJAE) to directly measure the quality of generated poses.

\subsubsection{Implementation Details}
Since PHOENIX14T lacks pose labels, we use OpenPose~\cite{Cao_Hidalgo_Simon_Wei_Sheikh_2021} to extract 2D joint coordinates and convert them to 3D using a skeletal correction model~\cite{Zelinka_Kanis_2020} as target poses. The Transformer-based Gloss Encoder is built with 2 layers, 4 heads, and an embedding size of 512. In addition, we set the timesteps $t$ of the diffusion model to 1000 and the number of inferences $i$ to 5. During training, we use the Adam optimizer~\cite{Kingma_Ba_2014} and a learning rate of $1 \times 10^{-3}$. Experiments are conducted using PyTorch on NVIDIA GeForce RTX 2080 Ti GPUs.

\subsection{Comparison with State-of-the-Arts}
\subsubsection{PHOENIX14T} Table~\ref{tab:main1} provides comparison results of the proposed Sign-IDD with other SOTA methods on PHOENIX14T. As shown in this table, Sign-IDD significantly outperforms other non-diffusion-based approaches, achieving 25.40\% and 24.80\% BLEU-1 on the DEV and TEST sets, respectively. Even compared with the best-performing non-diffusion-based method, our method achieves significant performance improvements on ROUGE and WER, \emph{e.g.}, Sign-IDD surpasses GEN-OBT~\cite{tang2022gloss} by margins of 3.09\% and 5.12\% on the TEST set. Considering the strong advantages of the diffusion model itself in content generation, we specifically compare our solution with several diffusion-based methods. It is noticeable that our method achieves higher BLEU than the most recent diffusion-based SLP method, \emph{e.g.}, Sign-IDD is 8.69\% and 1.58\% higher than G2P-DDM~\cite{xie2024g2p} on BLEU-1 and BLEU-4 metrics. 
Compared with diffusion-based methods using multiple hypothesis strategies during inference, e.g., D3DP-sign~\cite{shan2023diffusion} and GCDM~\cite{tang2024GCDM}, our method still achieves superior performance. 

\begin{figure*}[tbh]
  \centering
  \includegraphics[width=\textwidth]{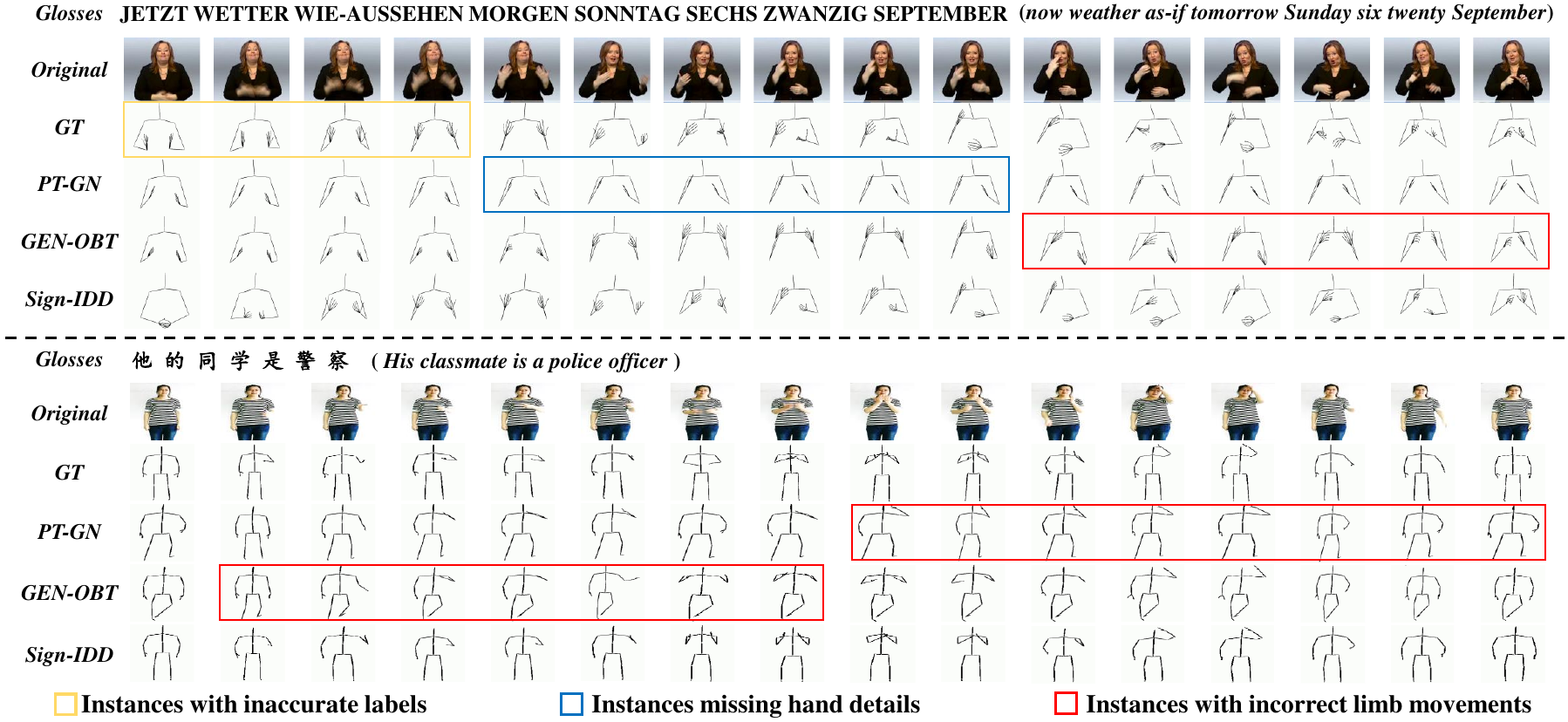}
  \vspace{-4mm}
  \caption{Visualization examples of generated poses on PHOENIX14T (top) and USTC-CSL (bottom). We compare Sign-IDD with PT-GN and GEN-OBT. Gloss annotations, original video frames, and ground-truth poses are attached for clear evaluation.}
  \vspace{-3mm}
  \label{fig:Visual 1}
\end{figure*}

\subsubsection{USTC-CSL} Table~\ref{tab:main2} shows comparison results on a challenging Chinese sign language benchmark USTC-CSL. Note that no existing work has reported the SLP performance on USTC-CSL, so we test several typical solutions under a fair setting. 
Sign-IDD achieves the best performance on all back-translation metrics, \emph{e.g.}, 65.26\% and 50.15\% on BLEU-1 and WER, which indicates that Sign-IDD can maintain high semantic accuracy during sign generation. In addition, our performance on direct metrics outperforms previous non-diffusion-based solutions and is especially significantly better than diffusion-based D3DP-sign~\cite{shan2023diffusion} by 7.07\% on MPJPE. This further demonstrates the superiority of the poses generated by Sign-IDD in terms of the accuracy of joint positions and bone orientations. 

\subsubsection{Visualization Results} In Figure~\ref{fig:Visual 1}, we visualize sign pose examples generated by the proposed Sign-IDD and other methods, \emph{i.e.}, PT-GN~\cite{saunders2020progressive} and GEN-OBT~\cite{tang2022gloss}.
In the top example, the sign poses generated by Sign-IDD are noticeably superior to PT-GN, particularly in hand details (blue box), and demonstrate more accurate limb movements compared to GEN-OBT (red box). Furthermore, even in cases where the ground truth provides inaccurate pose labels due to motion blur (yellow box), Sign-IDD consistently generates clear and precise results. The bottom example shows that Sign-IDD outperforms existing methods in generating both upper and lower limb movements, further highlighting the advantages of our approach in accurately generating joint positions and bone orientations during sign production.

\begin{table}[tbp]
\renewcommand\arraystretch{1.1}
\caption{Ablation results of modules. \textbf{ACD}: Attribute Controllable Diffusion, \textbf{ID}: Iconicity Disentanglement.}
\vspace{-3mm}
\label{tab:ablation modules}
\centering
\resizebox{0.46\textwidth}{!}{
\begin{tabular}{ccccccccc}
\Xhline{1pt}
   \multicolumn{1}{l}{\multirow{2}{*}{Methods}} & \multicolumn{3}{c}{DEV} & ~ & \multicolumn{3}{c}{TEST} \\
   \cline{2-4}\cline{6-8}
   \multicolumn{1}{c}{} & B1{$\uparrow$} & B4{$\uparrow$} & WER{$\downarrow$} & ~ & B1{$\uparrow$} & B4{$\uparrow$} & WER{$\downarrow$}\\
\Xhline{0.5pt}
   \multicolumn{1}{l}{Base}  &21.38 &7.06  &85.29  & ~ &21.50 &7.11  &84.74 \\
   \multicolumn{1}{l}{Base+ID}  &24.97 &8.51 &78.52 & ~ &23.46 &8.21 &77.53\\
    \multicolumn{1}{l}{Base+ACD}  &24.33  &8.40  &82.04  & ~ &23.42  &8.16  &79.92 \\
    \Xhline{0.5pt}
     \multicolumn{1}{l}{Base+ID+ACD}  &\bf{25.40}  &\bf{8.93}  &\bf{77.72}  & ~ &\bf{24.80}  &\bf{9.08}  &\bf{76.66} \\
\Xhline{1pt}
\end{tabular}}
\vspace{-3mm}
\end{table}

\begin{figure}[!htb]
\centering
\includegraphics[width=0.98\columnwidth]{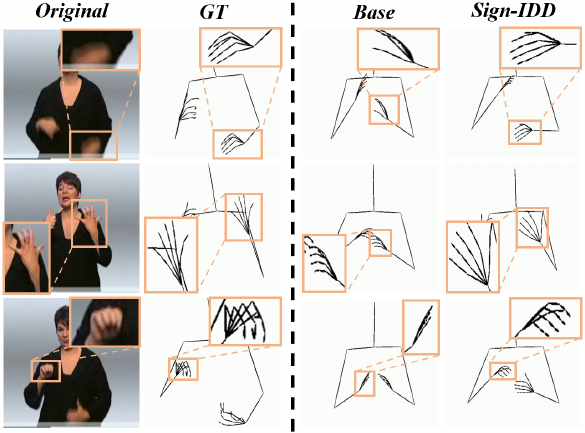}
\caption{Visualization results of Sign-IDD and Base, which is a diffusion-based baseline without ID and ACD modules.}
\vspace{-5mm}
\label{fig:Visual 3}
\end{figure}

\begin{table}[tbp]
\renewcommand\arraystretch{1.1}
\caption{Ablation results of parameters on PHOENIX14T.}
\vspace{-3mm}
\label{tab:ablation parameters}
\centering
\resizebox{0.45\textwidth}{!}{
\begin{tabular}{cccccccccc}
\Xhline{1pt}
   \multicolumn{2}{c}{\multirow{2}{*}{Methods}} & \multicolumn{3}{c}{DEV} & ~ & \multicolumn{3}{c}{TEST} \\
   \cline{3-5}\cline{7-9}
   \multicolumn{2}{c}{} & B1{$\uparrow$} & B4{$\uparrow$} & WER{$\downarrow$} & ~ & B1{$\uparrow$} & B4{$\uparrow$} & WER{$\downarrow$}\\
\Xhline{0.5pt}
   \multicolumn{1}{c|}{\multirow{3}{*}{$\lambda$}} & 0.01 &21.34  &6.93  &85.53  & ~ &20.89  &6.97  &84.43 \\
   \multicolumn{1}{c|}{} & 0.1 & \bf{25.40} & \bf{8.93} & \bf{77.72} & ~ & \bf{24.80} & \bf{9.08} & \bf{76.66}\\
    \multicolumn{1}{c|}{} & 1 &20.85  &6.85  &88.66  & ~ &19.95  &6.46  &88.38 \\
\Xhline{0.5pt}
   \multicolumn{1}{c|}{\multirow{3}{*}{$t$}} & 500 & 24.11 & 8.07 & 81.56 & ~ & 23.63 & 8.43 & 79.90\\
   \multicolumn{1}{c|}{} & 1000 & \bf{25.40} & \bf{8.93} & \bf{77.72} & ~ & \bf{24.80} & \bf{9.08} & \bf{76.66}\\
   \multicolumn{1}{c|}{} & 1500 & 23.09 & 7.96 & 83.88 & ~ & 22.77 & 7.53 & 83.99\\
\Xhline{0.5pt}
   \multicolumn{1}{c|}{\multirow{3}{*}{$i$}} & 1 & 24.05 & 8.38 & 82.39 & ~ & 23.60 & 8.42 & 81.73\\
   \multicolumn{1}{c|}{} & 5 & \bf{25.40} & \bf{8.93} & \bf{77.72} & ~ & \bf{24.80} & \bf{9.08} & \bf{76.66}\\
   \multicolumn{1}{c|}{} & 10 &20.85  &6.85  &88.66  & ~ &19.95  &6.46  &88.38 \\
\Xhline{1pt}
\end{tabular}}
\vspace{-5mm}
\end{table}

\subsection{Ablation Study}
In this subsection, we present ablation results to verify the effectiveness of the Sign-IDD. All results are evaluated on PHOENIX14T, while USTC-CSL is not used for ablations. 

\subsubsection{Retain Stronger Sign Semantics}
Table~\ref{tab:ablation modules} shows the ablation results of ID and ACD modules. We set a diffusion-based SLP model as a baseline, denoted as Base, which only uses 3D joint coordinates and replaces the proposed ACD with a denoiser that removes attribute separation/control layers. Base+ID refers to incorporating ID into Base (\emph{i.e.}, introducing iconicity disentanglement strategy), which improves BLEU-1/BLEU-4 by 1.96\%/1.10\% on TEST. In addition, introducing ACD with an attribute control mechanism based on Base also improves performance, and the WER reaches 82.04\%/79.92\% (compared to 85.29\%/84.74\%). When both ID and ACD modules are adopted, our method achieves the best performance on all back-translation metrics, which indicates that poses generated by Sign-IDD retain richer and more accurate sign semantics.

\subsubsection{Generate Clearer Local Details}
In Figure~\ref{fig:Visual 3}, the visualization further shows the direct impact of the proposed ID and ACD modules on the generated sign poses. In the top and bottom examples, the Base generates overlapping fingers, obscuring important hand movements. In contrast, Sign-IDD generates clearer local details and the posture of fingers is also closer to the truth. 

\subsubsection{Capture More Accurate Limb Orientation}
In the middle example of Figure~\ref{fig:Visual 3}, although Base correctly captures the relative positions of the hand joints, the finger bone orientations differ significantly from the ground truth. In contrast, Sign-IDD accurately aligns both joint positions and bone orientations, with all fingers pointing in the correct direction, closely matching the target pose. Additionally, the top example further demonstrates that incorporating iconicity disentanglement strategy and attribute controllable diffusion enhances the orientation accuracy of generated limbs.

\subsubsection{Analysis of Parameters} Table~\ref{tab:ablation parameters} presents the ablation results for Sign-IDD, focusing on the weight $\lambda$ of the $\mathcal{L}_{bone}$, the time step $t$ in diffusion, and the sampling step $i$ during inference. The model achieves optimal performance with $\lambda=0.1$, $t=1000$, and $angi=5$.

\section{Conclusions}
This work proposes to enhance the accuracy and naturalness of generated sign poses by modeling both positions and associations of joints. We introduce a novel iconicity disentanglement strategy that transforms 3D joint representations into 4D bone representations, covering 3D orientation and 1D length. We further design an attribute controllable diffusion module to separate skeletal attributes and utilize them to guide pose generation. Extensive experiments on two benchmarks validate the effectiveness of our approach.

\section{Acknowledgments}
This work was supported by the National Natural Science Foundation of China (Grants No. U23B2031, 61932009, U20A20183, 62272144, 62302141, 62331003), the Anhui Provincial Natural Science Foundation, China (Grant No. 2408085QF191), the Major Project of Anhui Province (Grant No. 202423k09020001), and the Fundamental Research Funds for the Central Universities (Grants No. JZ2024HGTA0178, JZ2024HGTB0255).

\bibliography{aaai25}

\begin{thebibliography}{54}
\providecommand{\natexlab}[1]{#1}

\bibitem[{Baltatzis et~al.(2024)Baltatzis, Potamias, Ververas, Sun, Deng, and
  Zafeiriou}]{Baltatzis_Potamias_Ververas_Sun_Deng_Zafeiriou_2023}
Baltatzis, V.; Potamias, R.~A.; Ververas, E.; Sun, G.; Deng, J.; and Zafeiriou,
  S. 2024.
\newblock Neural Sign Actors: A Diffusion Model for 3D Sign Language Production
  from Text.
\newblock In \emph{IEEE/CVF Conference on Computer Vision and Pattern
  Recognition}, 1985--1995.

\bibitem[{Cai et~al.(2024)Cai, Hu, Hou, Yao, and Huang}]{cai2024disentangled}
Cai, Q.; Hu, X.; Hou, S.; Yao, L.; and Huang, Y. 2024.
\newblock Disentangled Diffusion-Based 3D Human Pose Estimation with
  Hierarchical Spatial and Temporal Denoiser.
\newblock In \emph{AAAI Conference on Artificial Intelligence}, 882--890.

\bibitem[{Camgoz et~al.(2018)Camgoz, Hadfield, Koller, Ney, and
  Bowden}]{Camgoz_Hadfield_Koller_Ney_Bowden_2018}
Camgoz, N.~C.; Hadfield, S.; Koller, O.; Ney, H.; and Bowden, R. 2018.
\newblock Neural Sign Language Translation.
\newblock In \emph{IEEE/CVF Conference on Computer Vision and Pattern
  Recognition}, 7784--7793.

\bibitem[{Camgoz et~al.(2020)Camgoz, Koller, Hadfield, and
  Bowden}]{Cihan_Camgoz_Koller_Hadfield_Bowden_2020}
Camgoz, N.~C.; Koller, O.; Hadfield, S.; and Bowden, R. 2020.
\newblock Sign Language Transformers: Joint End-to-End Sign Language
  Recognition and Translation.
\newblock In \emph{IEEE/CVF Conference on Computer Vision and Pattern
  Recognition}, 10023--10033.

\bibitem[{Cao et~al.(2017)Cao, Simon, Wei, and
  Sheikh}]{Cao_Hidalgo_Simon_Wei_Sheikh_2021}
Cao, Z.; Simon, T.; Wei, S.-E.; and Sheikh, Y. 2017.
\newblock Realtime Multi-Person 2d Pose Estimation Using Part Affinity Fields.
\newblock In \emph{IEEE/CVF Conference on Computer Vision and Pattern
  Recognition}, 7291--7299.

\bibitem[{Cui, Liu, and Zhang(2017)}]{Cui_Liu_Zhang_2017}
Cui, R.; Liu, H.; and Zhang, C. 2017.
\newblock Recurrent Convolutional Neural Networks for Continuous Sign Language
  Recognition by Staged Optimization.
\newblock In \emph{IEEE/CVF Conference on Computer Vision and Pattern
  Recognition}, 7361--7369.

\bibitem[{Fang et~al.(2023)Fang, Sui, Zhang, and Tian}]{fang2023signdiff}
Fang, S.; Sui, C.; Zhang, X.; and Tian, Y. 2023.
\newblock SignDiff: Learning Diffusion Models for American Sign Language
  Production.
\newblock \emph{arXiv Preprint ArXiv:2308.16082}, arXiv--2308.

\bibitem[{Goodfellow et~al.(2014)Goodfellow, Pouget-Abadie, Mirza, Xu,
  Warde-Farley, Ozair, Courville, and
  Bengio}]{Goodfellow_Pouget-Abadie_Mirza_Xu_Warde-Farley_Ozair_Courville_Bengio_2017}
Goodfellow, I.; Pouget-Abadie, J.; Mirza, M.; Xu, B.; Warde-Farley, D.; Ozair,
  S.; Courville, A.; and Bengio, Y. 2014.
\newblock Generative Adversarial Nets.
\newblock \emph{Advances in Neural Information Processing Systems},
  2672–2680.

\bibitem[{Guo et~al.(2021)Guo, Tang, Hong, and Wang}]{guo2021sign}
Guo, D.; Tang, S.; Hong, R.; and Wang, M. 2021.
\newblock Sign Language Recognition.
\newblock \emph{Multimedia for Accessible Human Computer Interfaces}, 23--59.

\bibitem[{Guo, Tang, and Wang(2019)}]{Guo_Tang_Wang}
Guo, D.; Tang, S.; and Wang, M. 2019.
\newblock Connectionist Temporal Modeling of Video and Language: A Joint Model
  for Translation and Sign Labeling.
\newblock In \emph{International Joint Conference on Artificial Intelligence},
  751--757.

\bibitem[{Guo et~al.(2017)Guo, Zhou, Li, and Wang}]{Guo_Zhou_Li_Wang_2018}
Guo, D.; Zhou, W.; Li, H.; and Wang, M. 2017.
\newblock Online Early-Late Fusion Based on Adaptive HMM for Sign Language
  Recognition.
\newblock \emph{ACM Transactions on Multimedia Computing, Communications, and
  Applications}, 1--18.

\bibitem[{Guo et~al.(2018)Guo, Zhou, Li, and Wang}]{guo2018hierarchical}
Guo, D.; Zhou, W.; Li, H.; and Wang, M. 2018.
\newblock Hierarchical LSTM for Sign Language Translation.
\newblock In \emph{AAAI Conference on Artificial Intelligence}, 6845--6852.

\bibitem[{Guo et~al.(2024)Guo, He, Tang, Wang, and Cheng}]{guo2024shaping}
Guo, M.; He, J.; Tang, S.; Wang, Z.; and Cheng, L. 2024.
\newblock Shaping a Stabilized Video by Mitigating Unintended Changes for
  Concept-Augmented Video Editing.
\newblock \emph{arXiv preprint arXiv:2410.12526}.

\bibitem[{Ho, Jain, and Abbeel(2020)}]{Ho_Jain_Abbeel_Berkeley}
Ho, J.; Jain, A.; and Abbeel, P. 2020.
\newblock Denoising Diffusion Probabilistic Models.
\newblock \emph{Advances in Neural Information Processing Systems}, 6840--6851.

\bibitem[{Ho et~al.(2022)Ho, Salimans, Gritsenko, Chan, Norouzi, and
  Fleet}]{Ho_Salimans_Gritsenko_Chan_Norouzi_Fleet_2022}
Ho, J.; Salimans, T.; Gritsenko, A.; Chan, W.; Norouzi, M.; and Fleet, D.~J.
  2022.
\newblock Video Diffusion Models.
\newblock \emph{Advances in Neural Information Processing Systems}, 8633--8646.

\bibitem[{Huang et~al.(2018)Huang, Zhou, Zhang, Li, and Li}]{huang2018video}
Huang, J.; Zhou, W.; Zhang, Q.; Li, H.; and Li, W. 2018.
\newblock Video-Based Sign Language Recognition without Temporal Segmentation.
\newblock In \emph{AAAI Conference on Artificial Intelligence}, 2257--2264.

\bibitem[{Huang et~al.(2022)Huang, Lam, Wang, Su, Yu, Ren, and
  Zhao}]{huang2022fastdiff}
Huang, R.; Lam, M.; Wang, J.; Su, D.; Yu, D.; Ren, Y.; and Zhao, Z. 2022.
\newblock FastDiff: A Fast Conditional Diffusion Model for High-Quality Speech
  Synthesis.
\newblock In \emph{International Joint Conference on Artificial Intelligence},
  4157--4163.

\bibitem[{Huang et~al.(2021)Huang, Pan, Zhao, and Tian}]{huang2021towards}
Huang, W.; Pan, W.; Zhao, Z.; and Tian, Q. 2021.
\newblock Towards Fast and High-Quality Sign Language Production.
\newblock In \emph{ACM International Conference on Multimedia}, 3172--3181.

\bibitem[{Kingma and Ba(2015)}]{Kingma_Ba_2014}
Kingma, D.~P.; and Ba, J. 2015.
\newblock Adam: A Method for Stochastic Optimization.
\newblock In \emph{International Conference on Learning Representations},
  1--15.

\bibitem[{Kingma and Welling(2014)}]{kingma2013auto}
Kingma, D.~P.; and Welling, M. 2014.
\newblock Auto-Encoding Variational Bayes.
\newblock \emph{Stat}, 1.

\bibitem[{Koller(2020)}]{Koller_2020}
Koller, O. 2020.
\newblock Quantitative Survey of the State of the Art in Sign Language
  Recognition.
\newblock \emph{arXiv Preprint ArXiv:2008.09918}, arXiv--2008.

\bibitem[{Li et~al.(2024{\natexlab{a}})Li, Cong, Wu, Bai, Wang, and
  Zhao}]{li2024srconvnet}
Li, F.; Cong, R.; Wu, J.; Bai, H.; Wang, M.; and Zhao, Y. 2024{\natexlab{a}}.
\newblock SRConvNet: A Transformer-Style ConvNet for Lightweight Image
  Super-Resolution.
\newblock \emph{International Journal of Computer Vision}, 1--17.

\bibitem[{Li et~al.(2024{\natexlab{b}})Li, Wu, Li, Bai, Cong, and
  Zhao}]{li2024enhanced}
Li, F.; Wu, Y.; Li, A.; Bai, H.; Cong, R.; and Zhao, Y. 2024{\natexlab{b}}.
\newblock Enhanced Video Super-Resolution Network towards Compressed Data.
\newblock \emph{ACM Transactions on Multimedia Computing, Communications and
  Applications}, 1--21.

\bibitem[{Li et~al.(2022)Li, Thickstun, Gulrajani, Liang, and
  Hashimoto}]{Li_Thickstun_Gulrajani_Liang_Hashimoto_2022}
Li, X.; Thickstun, J.; Gulrajani, I.; Liang, P.~S.; and Hashimoto, T.~B. 2022.
\newblock Diffusion-LM Improves Controllable Text Generation.
\newblock \emph{Advances in Neural Information Processing Systems}, 4328--4343.

\bibitem[{Makhzani et~al.(2015)Makhzani, Shlens, Jaitly, Goodfellow, and
  Frey}]{makhzani2015adversarial}
Makhzani, A.; Shlens, J.; Jaitly, N.; Goodfellow, I.; and Frey, B. 2015.
\newblock Adversarial Autoencoders.
\newblock \emph{arXiv e-prints}, arXiv--1511.

\bibitem[{Mazumder et~al.(2021)Mazumder, Mukhopadhyay, Namboodiri, and
  Jawahar}]{Mazumder_Mukhopadhyay_Namboodiri_Jawahar_2021}
Mazumder, S.; Mukhopadhyay, R.; Namboodiri, V.~P.; and Jawahar, C. 2021.
\newblock Translating Sign Language Videos to Talking Faces.
\newblock In \emph{Indian Conference on Computer Vision, Graphics and Image
  Processing}, 1--10.

\bibitem[{McDonald et~al.(2016)McDonald, Wolfe, Schnepp, Hochgesang, Jamrozik,
  Stumbo, Berke, Bialek, and
  Thomas}]{McDonald_Wolfe_Schnepp_Hochgesang_Jamrozik_Stumbo_Berke_Bialek_Thomas_2016}
McDonald, J.; Wolfe, R.; Schnepp, J.; Hochgesang, J.; Jamrozik, D.~G.; Stumbo,
  M.; Berke, L.; Bialek, M.; and Thomas, F. 2016.
\newblock An Automated Technique for Real-Time Production of Lifelike
  Animations of American Sign Language.
\newblock \emph{Universal Access in the Information Society}, 551--566.

\bibitem[{Moryossef et~al.(2021)Moryossef, Yin, Neubig, and
  Goldberg}]{moryossef2021data}
Moryossef, A.; Yin, K.; Neubig, G.; and Goldberg, Y. 2021.
\newblock Data Augmentation for Sign Language Gloss Translation.
\newblock In \emph{International Workshop on Automatic Translation for Signed
  and Spoken Languages}, 1--11.

\bibitem[{Nielsen and Dingemanse(2021)}]{nielsen2021iconicity}
Nielsen, A.~K.; and Dingemanse, M. 2021.
\newblock Iconicity in Word Learning and Beyond: A Critical Review.
\newblock \emph{Language and Speech}, 52--72.

\bibitem[{Orbay and Akarun(2020)}]{Orbay_Akarun_2020}
Orbay, A.; and Akarun, L. 2020.
\newblock Neural Sign Language Translation by Learning Tokenization.
\newblock In \emph{IEEE International Conference on Automatic Face and Gesture
  Recognition}, 222--228.

\bibitem[{Othman and Jemni(2011)}]{othman2011statistical}
Othman, A.; and Jemni, M. 2011.
\newblock Statistical Sign Language Machine Translation: from English Written
  Text to American Sign Language Gloss.
\newblock \emph{International Journal of Computer Science Issues}, 65.

\bibitem[{Saunders, Camgoz, and
  Bowden(2020{\natexlab{a}})}]{Saunders_Camgoz_Bowden_2020_A}
Saunders, B.; Camgoz, N.~C.; and Bowden, R. 2020{\natexlab{a}}.
\newblock Adversarial Training for Multi-Channel Sign Language Production.
\newblock In \emph{British Machine Vision Conference}, 1--15.

\bibitem[{Saunders, Camgoz, and
  Bowden(2020{\natexlab{b}})}]{saunders2020progressive}
Saunders, B.; Camgoz, N.~C.; and Bowden, R. 2020{\natexlab{b}}.
\newblock Progressive Transformers for End-to-End Sign Language Production.
\newblock In \emph{European Conference on Computer Vision}, 687--705.

\bibitem[{Saunders, Camgoz, and
  Bowden(2021{\natexlab{a}})}]{saunders2021continuous}
Saunders, B.; Camgoz, N.~C.; and Bowden, R. 2021{\natexlab{a}}.
\newblock Continuous 3d Multi-Channel Sign Language Production via Progressive
  Transformers and Mixture Density Networks.
\newblock \emph{International journal of computer vision}, 2113--2135.

\bibitem[{Saunders, Camgoz, and Bowden(2021{\natexlab{b}})}]{saunders2021mixed}
Saunders, B.; Camgoz, N.~C.; and Bowden, R. 2021{\natexlab{b}}.
\newblock Mixed Signals: Sign Language Production via A Mixture of Motion
  Primitives.
\newblock In \emph{IEEE/CVF International Conference on Computer Vision},
  1919--1929.

\bibitem[{Saunders, Camgoz, and
  Bowden(2022{\natexlab{a}})}]{saunders2022signing}
Saunders, B.; Camgoz, N.~C.; and Bowden, R. 2022{\natexlab{a}}.
\newblock Signing at Scale: Learning to Co-Articulate Signs for Large-Scale
  Photo-Realistic Sign Language Production.
\newblock In \emph{IEEE/CVF Conference on Computer Vision and Pattern
  Recognition}, 5141--5151.

\bibitem[{Saunders, Camgoz, and
  Bowden(2022{\natexlab{b}})}]{saunders2022skeletal}
Saunders, B.; Camgoz, N.~C.; and Bowden, R. 2022{\natexlab{b}}.
\newblock Skeletal Graph Self-Attention: Embedding a Skeleton Inductive Bias
  into Sign Language Production.
\newblock In \emph{International Workshop on Sign Language Translation and
  Avatar Technology}, 95--102.

\bibitem[{Segouat(2009)}]{Segouat_2009}
Segouat, J. 2009.
\newblock A Study of Sign Language Coarticulation.
\newblock \emph{ACM Sigaccess Accessibility and Computing}, 31--38.

\bibitem[{Shan et~al.(2023)Shan, Liu, Zhang, Wang, Han, Wang, Ma, and
  Gao}]{shan2023diffusion}
Shan, W.; Liu, Z.; Zhang, X.; Wang, Z.; Han, K.; Wang, S.; Ma, S.; and Gao, W.
  2023.
\newblock Diffusion-Based 3D Human Pose Estimation with Multi-Hypothesis
  Aggregation.
\newblock In \emph{IEEE/CVF International Conference on Computer Vision},
  14761--14771.

\bibitem[{Shang et~al.(2024)Shang, Shan, Liu, Wang, Wang, Zhang, and
  Zhang}]{shang2024resdiff}
Shang, S.; Shan, Z.; Liu, G.; Wang, L.; Wang, X.; Zhang, Z.; and Zhang, J.
  2024.
\newblock Resdiff: Combining Cnn and Diffusion Model for Image
  Super-Resolution.
\newblock In \emph{AAAI Conference on Artificial Intelligence}, 8975--8983.

\bibitem[{Sohl-Dickstein et~al.(2015)Sohl-Dickstein, Weiss, Maheswaranathan,
  and Ganguli}]{Sohl-Dickstein_Weiss_Maheswaranathan_Ganguli_2015}
Sohl-Dickstein, J.; Weiss, E.; Maheswaranathan, N.; and Ganguli, S. 2015.
\newblock Deep Unsupervised Learning Using Nonequilibrium Thermodynamics.
\newblock In \emph{International Conference on Machine Learning}, 2256--2265.

\bibitem[{Song et~al.(2024)Song, Guo, Yang, Tang, and Wang}]{song2024emotional}
Song, P.; Guo, D.; Yang, X.; Tang, S.; and Wang, M. 2024.
\newblock Emotional Video Captioning With Vision-Based Emotion Interpretation
  Network.
\newblock \emph{IEEE Transactions on Image Processing}, 1122--1135.

\bibitem[{Song et~al.(2023)Song, Guo, Yang, Tang, Yang, and
  Wang}]{song2023emotion}
Song, P.; Guo, D.; Yang, X.; Tang, S.; Yang, E.; and Wang, M. 2023.
\newblock Emotion-Prior Awareness Network for Emotional Video Captioning.
\newblock In \emph{ACM International Conference on Multimedia}, 589--600.

\bibitem[{Stoll et~al.(2020)Stoll, Camgoz, Hadfield, and
  Bowden}]{Stoll_Camgoz_Hadfield_Bowden_2020}
Stoll, S.; Camgoz, N.~C.; Hadfield, S.; and Bowden, R. 2020.
\newblock Text2Sign: Towards Sign Language Production Using Neural Machine
  Translation and Generative Adversarial Networks.
\newblock \emph{International Journal of Computer Vision}, 891--908.

\bibitem[{Tang et~al.(2022{\natexlab{a}})Tang, Guo, Hong, and
  Wang}]{tang2022graph}
Tang, S.; Guo, D.; Hong, R.; and Wang, M. 2022{\natexlab{a}}.
\newblock Graph-Based Multimodal Sequential Embedding for Sign Language
  Translation.
\newblock \emph{IEEE Transactions on Multimedia}, 4433--4445.

\bibitem[{Tang et~al.(2022{\natexlab{b}})Tang, Hong, Guo, and
  Wang}]{tang2022gloss}
Tang, S.; Hong, R.; Guo, D.; and Wang, M. 2022{\natexlab{b}}.
\newblock Gloss Semantic-Enhanced Network with Online Back-Translation for Sign
  Language Production.
\newblock In \emph{ACM International Conference on Multimedia}, 5630--5638.

\bibitem[{Tang et~al.(2024)Tang, Xue, Wu, Wang, and Hong}]{tang2024GCDM}
Tang, S.; Xue, F.; Wu, J.; Wang, S.; and Hong, R. 2024.
\newblock Gloss-Driven Conditional Diffusion Models for Sign Language
  Production.
\newblock \emph{ACM Transactions on Multimedia Computing, Communications, and
  Applications}.

\bibitem[{Vaswani et~al.(2017)Vaswani, Shazeer, Parmar, Uszkoreit, Jones,
  Gomez, Kaiser, and Polosukhin}]{vaswani2017attention}
Vaswani, A.; Shazeer, N.; Parmar, N.; Uszkoreit, J.; Jones, L.; Gomez, A.~N.;
  Kaiser, L.; and Polosukhin, I. 2017.
\newblock Attention is All You Need.
\newblock In \emph{Advances in Neural Information Processing Systems},
  6000--6010.

\bibitem[{Viegas et~al.(2023)Viegas, Inan, Quandt, and
  Alikhani}]{Viegas_Inan_Quandt_Alikhani}
Viegas, C.; Inan, M.; Quandt, L.; and Alikhani, M. 2023.
\newblock Including Facial Expressions in Contextual Embeddings for Sign
  Language Generation.
\newblock In \emph{Joint Conference on Lexical and Computational Semantics},
  1--10.

\bibitem[{Wang, Chai, and Chen(2019)}]{Wang_Chai_Chen_2019}
Wang, H.; Chai, X.; and Chen, X. 2019.
\newblock A Novel Sign Language Recognition Framework Using Hierarchical
  Grassmann Covariance Matrix.
\newblock \emph{IEEE Transactions on Multimedia}, 2806--2814.

\bibitem[{Wei et~al.(2024)Wei, Zhang, Li, Wang, Tang, and
  Zhang}]{wei2024leveraging}
Wei, Y.; Zhang, Y.; Li, K.; Wang, F.; Tang, S.; and Zhang, Z. 2024.
\newblock Leveraging Vision-Language Prompts for Real-World Image Restoration
  and Enhancement.
\newblock \emph{Computer Vision and Image Understanding}, 104222.

\bibitem[{Wu, Hong, and Tang(2024)}]{wu2024intermediary}
Wu, J.; Hong, R.; and Tang, S. 2024.
\newblock Intermediary-Generated Bridge Network for RGB-D Cross-Modal
  Re-Identification.
\newblock \emph{ACM Transactions on Intelligent Systems and Technology}, 1--25.

\bibitem[{Xie et~al.(2024)Xie, Zhang, Taiying, Tang, Du, and Li}]{xie2024g2p}
Xie, P.; Zhang, Q.; Taiying, P.; Tang, H.; Du, Y.; and Li, Z. 2024.
\newblock G2P-DDM: Generating Sign Pose Sequence from Gloss Sequence with
  Discrete Diffusion Model.
\newblock In \emph{AAAI Conference on Artificial Intelligence}, 6234--6242.

\bibitem[{Zelinka and Kanis(2020)}]{Zelinka_Kanis_2020}
Zelinka, J.; and Kanis, J. 2020.
\newblock Neural Sign Language Synthesis: Words Are Our Glosses.
\newblock In \emph{IEEE/CVF Winter Conference on Applications of Computer
  Vision}, 3395--3403.

\end{thebibliography}

\end{document}